\pdfoutput=1
\documentclass{article}

 \usepackage[main, final]{neurips_2026}

\usepackage[utf8]{inputenc} 
\usepackage[T1]{fontenc}    
\usepackage{hyperref}       
\usepackage{url}            
\usepackage{booktabs}       
\usepackage{amsfonts}       
\usepackage{nicefrac}       
\usepackage{microtype}      
\usepackage{xcolor}         
\usepackage{amsmath}
\usepackage{graphicx}
\usepackage{multirow}

\title{Selection, Not Fusion: Radar-Modulated State Space Models for Radar-Camera Depth Estimation}

\author{%
  Zhangcheng Hou \\
  School of Science and Technology\\
  Keio University\\
  Yokohama 223-8522, Japan \\
  \texttt{zhangcheng@ohtsuki.ics.keio.ac.jp} \\
  \And
  Tomoaki Ohtsuki \\
  School of Science and Technology\\
  Keio University\\
  Yokohama 223-8522, Japan \\
  \texttt{ohtsuki@ics.keio.ac.jp} \\
}

\newcommand{\Dl}{\boldsymbol{\Delta}}
\newcommand{\FM}{FusionMamba$^\dagger$}

\begin{document}

\maketitle

\begin{abstract}
  Radar camera depth estimation turns a very sparse but metric radar signal into a dense depth map. Existing methods combine radar and image features outside the sequence operator of the backbone, through concatenation, confidence gating, sparse supervision, or graph based extraction. Mamba variants for multimodal fusion also leave the selection mechanism itself unimodal. We argue that selection is the right place for radar to enter. We introduce Radar-Modulated Selection (RMS), which injects radar inside the selective scan of Mamba. Radar adds learned perturbations, initialised at zero, to the step size $\Dl$ and the readout $\mathbf{C}$. The input projection $\mathbf{B}$ and the state matrix $\mathbf{A}$ never see radar. At initialisation the block equals a pretrained image Mamba, so radar gains influence only where it improves accuracy. The construction also couples the two modalities at linear cost and falls back to the image backbone when radar is missing. We deploy RMS in a Multi-View Scan Pyramid (MVSP) that matches the fusion operator to the reach of radar at each scale. On nuScenes, SemoDepth reduces MAE by $34.0\%$, $29.9\%$, and $29.9\%$ over the previous best within $50$, $70$, and $80$\,m. It also has the lowest single frame latency among methods with a public release ($26.8$\,ms). A feature blend placed before the scan adds no accuracy on top of RMS, which supports the view that selection inside the scan can replace fusion outside it. When twice as many spurious radar returns as true ones are injected, the MAE@$80$ of SemoDepth rises by $1.4\%$, against $33.6\%$ for a Mamba baseline that fuses radar before the scan.
\end{abstract}

\section{Introduction}

Dense metric depth is foundational for autonomous driving~\cite{kitti,nuscenes} and mobile robotics. It supports obstacle avoidance and localisation~\cite{adolfsson2022lidar} as well as 3D object detection and perception in bird's eye view~\cite{rcbevdet,chu2025racformer}. Monocular cameras are cheap, but they recover depth only up to scale~\cite{depth_anything} and degrade in rain, fog, and low light. Light detection and ranging (LiDAR) resolves scale, yet its power and cost confine it to premium vehicles. Automotive millimetre wave radar with frequency modulated continuous waves (FMCW) at 77 to 81\,GHz bridges this gap. It is inexpensive, robust to weather, directly metric, and already common in modern vehicles~\cite{bilik2019rise,han20234d,nuscenes}. The signal is very limited, though. A nuScenes radar sweep holds only 40 to 100 returns in the field of view (FoV) of the front camera, three orders of magnitude fewer than LiDAR. The returns also carry angular and height ambiguity, which emerging 4D radar datasets~\cite{palffy,RadarCam-Depth} resolve only in part.

Radar camera depth completion~\cite{uhrig2017sparsity} has a decade of designs. They include concatenation~\cite{Lin,RC-PDA,R4Dyn,DORN,Singh}, confidence aware gating~\cite{CaFNet}, sparse supervision~\cite{LiSBD}, graph based structure extraction~\cite{TacoDepth}, and two stage refinement of metric scale~\cite{RadarCam-Depth}. In every case radar and image features meet at the feature level, outside the sequence operator of the backbone. In parallel, selective state space models such as Mamba~\cite{mamba,mamba2}, Vim~\cite{vim}, VMamba~\cite{vmamba}, and MambaDepth~\cite{mambadepth} have become an efficient alternative to Transformers in vision. Their core is a selection mechanism, in which a step size $\Dl$ and a readout $\mathbf{C}$ computed from each token decide what the token keeps and emits. Existing Mamba variants for multimodal fusion~\cite{fusionmamba,mambadfuse} stack the streams of the two modalities before a shared scan, so the selection itself stays unimodal.

We let radar act on selection itself. Radar-Modulated Selection (RMS) injects radar inside the selective scan of Mamba~\cite{mamba}, and a Multi-View Scan Pyramid (MVSP) matches the fusion operator to the reach of radar at each decoder resolution. The resulting model, SemoDepth, sets a new state of the art on nuScenes~\cite{nuscenes} for mean absolute error (MAE) and root mean square error (RMSE) at every evaluation range. It also has the best MAE on ZJU-4DRadarCam~\cite{RadarCam-Depth} among methods without a monocular depth prior. We further show that selection inside the scan can replace fusion outside it, since a feature blend before the scan adds no accuracy on top of RMS. Our contributions are threefold.
\begin{itemize}
    \item Radar-Modulated Selection (RMS). This mechanism injects radar inside the selective scan of Mamba. Radar adds learned modulations, initialised at zero, to two selection parameters. These are the step size $\Dl$, which sets the memory horizon of each token, and the readout $\mathbf{C}$, which reads out the state. The input projection $\mathbf{B}$ and the state matrix $\mathbf{A}$ never see radar. Feature level fusion operates on static representations. Selection inside the scan instead modulates the recurrence dynamics that govern long range propagation. Radar then shapes not only how local features are composed but also how memory evolves along the sequence. It changes how information is stored and propagated, not merely how it is combined. Three advantages follow that fusion outside the scan cannot offer. (i) Coupling at linear cost. Radar shapes the recurrence at every token at $O(L)$ cost, never the $O(L^2)$ cost of cross attention. (ii) An initialisation that preserves the loss. At step zero the block equals a vanilla Mamba scan, so the pretrained image solution is preserved exactly and radar gradients grow only where they improve accuracy. (iii) A fallback to the image. With $\mathbf{A}$ and $\mathbf{B}$ unimodal, the model reduces to image behaviour under sparse or missing radar without any change to the architecture.
    \item Multi-View Scan Pyramid (MVSP). This decoder with three tiers matches the fusion operator to the reach of radar at each resolution. The two coarsest levels use RMS over the full image in four scan directions, a linear cost approximation of the 2D receptive field of a Transformer. The middle level uses RMS in windows around radar returns, so scan capacity is spent only where radar supports the image. The two finest levels use feature wise linear modulation (FiLM) at constant cost. Radar stays active at every decoder level, and scan compute concentrates where radar evidence reaches furthest. MVSP lowers MAE@$80$ by $18\%$ compared with FiLM at every level, so the allocation carries accuracy and is not merely an efficiency measure.
    \item State of the art on nuScenes and evidence that selection can replace fusion. SemoDepth sets a new state of the art for MAE and RMSE at every evaluation range and lowers MAE by about $30\%$ over the previous best. Controlled ablations show that modulating $\Dl$ and $\mathbf{C}$ helps, modulating $\mathbf{B}$ does not improve MAE@$80$, and a blend before the scan adds nothing. A corruption study confirms the fallback property. With radar removed at test time, MAE@$80$ rises by only $3.3\%$.
\end{itemize}

\section{Related work}
\subsection{Radar camera depth estimation}
\label{sec:related-rcd}
Methods fall into two families. Two stage approaches such as RadarCam-Depth~\cite{RadarCam-Depth} adapt a pretrained monocular depth model and learn a correction of its metric scale from radar. The image backbone never sees radar. Single stage methods learn dense depth end to end from radar and image. They differ in how radar enters the image pipeline, usually through feature fusion before or after the backbone operator. Early work~\cite{Lin,RC-PDA,R4Dyn,DORN,Singh} concatenates rasterised radar depth with the image channels and lets the network learn the fusion. CaFNet~\cite{CaFNet} adds a confidence gate that suppresses uncertain returns before mixing, which addresses the height and azimuth ambiguity of radar. Li et al.~\cite{LiSBD} avoid densification and supervise only at pixels that radar observes. TacoDepth~\cite{TacoDepth} extracts radar structure with a graph network and attends to image features through a pyramid of flash attention centred on radar points.

\subsection{Selective state space models in vision}
\label{sec:related-ssm}
Selective state space models (SSMs), introduced by Mamba~\cite{mamba}, extend linear time sequence models~\cite{s4} by computing the recurrence parameters from each token. A selective gate controls the memory horizon of each token, and a readout projection decides what leaves the hidden state. Vim~\cite{vim}, VMamba~\cite{vmamba}, and MambaDepth~\cite{mambadepth} adopt this design as a Transformer replacement for vision. Multimodal variants such as FusionMamba~\cite{fusionmamba} and MambaDFuse~\cite{mambadfuse} stack or concatenate features before a shared Mamba block. Their selection parameters $\Dl$ and $\mathbf{C}$ come from one fused representation and are never conditioned on each modality separately. This limits cross modal control within the selective dynamics that govern long range dependencies. SemoDepth computes $\Dl$ and $\mathbf{C}$ from both modalities, while the token stream and the input projection $\mathbf{B}$ never see radar.

\section{SemoDepth}
\label{sec:method}

\begin{figure}[t]
  \centering
  \includegraphics[width=\linewidth]{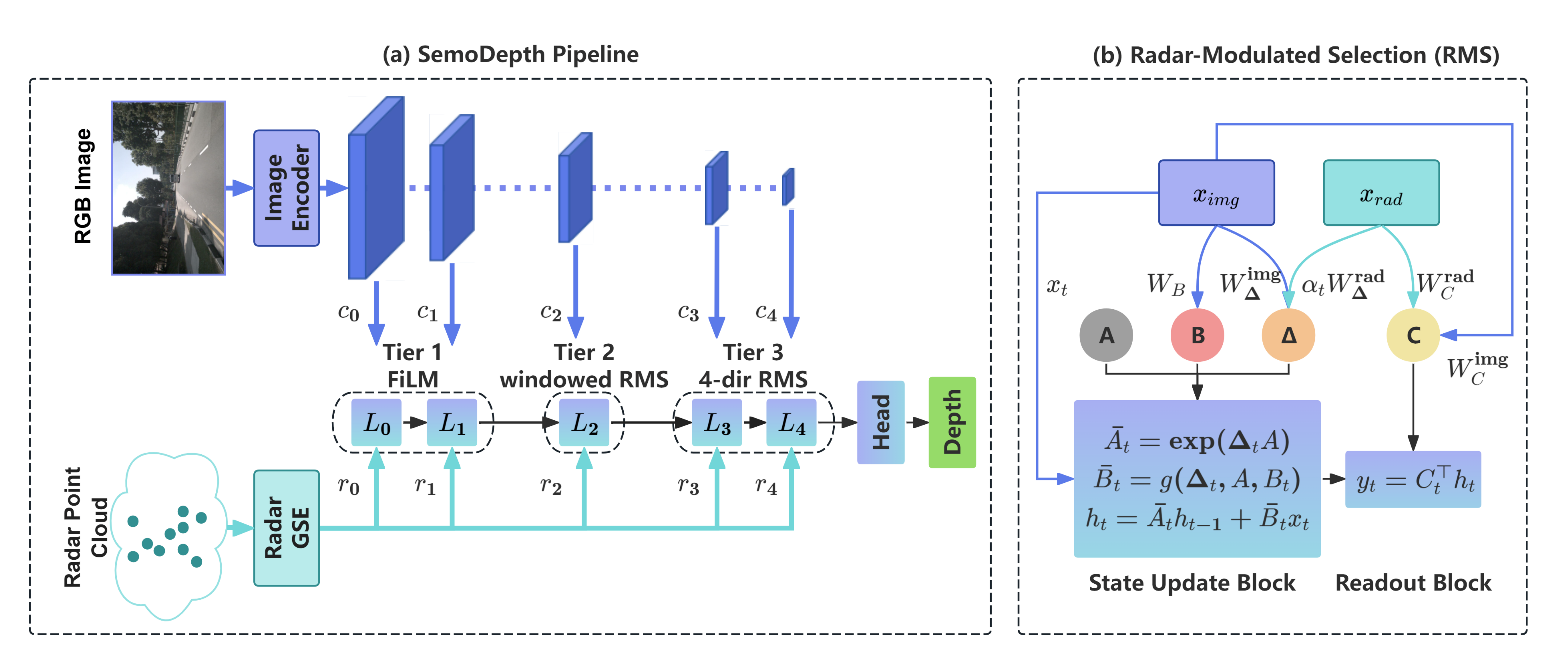}
  \caption{SemoDepth architecture. (a) Pipeline. A ResNet-34 image encoder and a radar graph structure extractor (GSE) built on PCA-GM produce the image pyramid $c_0, \dots, c_4$ and one radar feature map. Projections ($1{\times}1$) of this map at each level form the radar pyramid $r_0, \dots, r_4$. MVSP allocates fusion by resolution. FiLM modulates the two finest levels (Tier 1). Windowed RMS around each projected radar return serves the middle level (Tier 2). RMS over the full image in four directions serves the two coarsest levels (Tier 3). (b) Radar-Modulated Selection (RMS). Radar adds modulations to the step size $\Dl_t$ and readout $\mathbf{C}_t$ (teal arrows). The input projection $\mathbf{B}_t$ is computed from the image only (blue arrows), and the state matrix $\mathbf{A}$ never sees radar. The state update block applies zero order hold (ZOH) discretization and the recurrence $\mathbf{h}_t = \bar{\mathbf{A}}_t\mathbf{h}_{t-1} + \bar{\mathbf{B}}_t x_t$. The readout block emits $y_t = \mathbf{C}_t^{\!\top}\mathbf{h}_t$. All radar projections start at zero, so the block starts as vanilla Mamba.}
  \label{fig:arch}
\end{figure}

\subsection{Overview}
\label{sec:overview}

SemoDepth is a single stage depth estimator for radar and camera (Fig.~\ref{fig:arch}). A ResNet-34 encoder pretrained on ImageNet produces five feature maps $c_0, \dots, c_4$ at strides $1/2$ to $1/32$ of the input. In parallel, a radar graph structure extractor (GSE) that follows TacoDepth~\cite{TacoDepth} processes up to 512 radar returns per frame. Each return is described only by its geometry $(y/H,\,x/W,\,d/d_{\max})$, where $H$ and $W$ are the image height and width and $d$ is the metric range of the return. The GSE is a PCA-GM graph network with full adjacency~\cite{pcagm}, originally developed for deep graph matching, and its output is scattered into a radar feature map. Image and radar streams stay separate until the selective scan. The radar map is downsampled to each decoder resolution and passed through a $1{\times}1$ projection for that level, which yields the radar pyramid $r_0, \dots, r_4$. Decoder level $i$ fuses $c_i$ with $r_i$ through MVSP (\S\ref{sec:mvsp}). A regression head maps the finest decoder feature to metric depth at every pixel.

\subsection{Radar-Modulated Selection (RMS)}
\label{sec:rms}

We first describe RMS conceptually. The selective scan computes, for each token, a step size that controls how long the content of the token persists in the recurrence. It also computes a readout vector that controls what is exposed from the latent state. In a vanilla cross modal Mamba, both come from a single fused stream of image and radar. Radar then enters only as an additional input and never as a control over selection. RMS modulates the step size and the readout directly. At every token, radar contributes a learned correction to each, initialised at zero, while the input projection and the state matrix never see radar. The rest of this section makes the construction precise and defines all symbols of the scan in one place.

Let $\mathbf{x}_{\text{img},t}, \mathbf{x}_{\text{rad},t} \in \mathbb{R}^{D}$ be the image and radar tokens at position $t$ of a scanned sequence of length $L$ with $D$ channels. A vanilla Mamba block~\cite{mamba} runs a selective recurrence on the image stream, separately for each channel. A scalar input $x_t$, one channel of $\mathbf{x}_{\text{img},t}$, drives a hidden state $\mathbf{h}_t \in \mathbb{R}^{N}$ and produces a scalar output $y_t$,
\begin{equation}
\mathbf{h}_t = \bar{\mathbf{A}}_t\,\mathbf{h}_{t-1} + \bar{\mathbf{B}}_t\,x_t, \qquad y_t = \mathbf{C}_t^{\!\top}\mathbf{h}_t, \qquad t = 1, \dots, L.
\label{eq:ssm}
\end{equation}
ZOH discretization gives $\bar{\mathbf{A}}_t = \exp(\Dl_t \mathbf{A})$ and $\bar{\mathbf{B}}_t = (\Dl_t \mathbf{A})^{-1}\bigl(\exp(\Dl_t \mathbf{A}) - \mathbf{I}\bigr)\,\Dl_t \mathbf{B}_t$. Here $\mathbf{A} \in \mathbb{R}^{D \times N}$ is a trained matrix with negative entries, shared by all tokens, and $\mathbf{I}$ is the identity. Three quantities depend on the token. The step size $\Dl_t \in \mathbb{R}^D$ sets the memory horizon. The input projection $\mathbf{B}_t \in \mathbb{R}^N$ controls how the input enters the state. The readout $\mathbf{C}_t \in \mathbb{R}^N$ selects which part of the state is emitted. Vanilla Mamba computes all three from the image stream with trained linear maps,
\begin{equation}
\Dl_t = \mathrm{softplus}\bigl(\mathbf{W}_{\!\Delta}^{\text{img}}\,\mathbf{x}_{\text{img},t}\bigr), \qquad \mathbf{B}_t = \mathbf{W}_{\!B}\,\mathbf{x}_{\text{img},t}, \qquad \mathbf{C}_t = \mathbf{W}_{\!C}^{\text{img}}\,\mathbf{x}_{\text{img},t}.
\label{eq:sel-img}
\end{equation}
From a control theoretic view, these three parameters play distinct roles. $\mathbf{B}_t$ governs controllability, which is how inputs enter the system. $\mathbf{C}_t$ governs observability, which is the part of the hidden state that is exposed. $\Dl_t$ sets the memory timescale of the recurrence. RMS modulates $\Dl_t$ and $\mathbf{C}_t$ and keeps $\mathbf{B}_t$ driven by the image. This preserves the semantic content of image features. Radar influences temporal credit assignment through $\Dl_t$ and observation through $\mathbf{C}_t$, the two basic degrees of freedom over selection.

\subsubsection{Horizon modulation}
\label{sec:horizon}

The horizon pathway adds a gated radar term to the step size before the softplus,
\begin{equation}
\Dl_t = \mathrm{softplus}\bigl(\mathbf{W}_{\!\Delta}^{\text{img}}\,\mathbf{x}_{\text{img},t} + \alpha_t\,\mathbf{W}_{\!\Delta}^{\text{rad}}\,\mathbf{x}_{\text{rad},t}\bigr), \qquad \alpha_t = \sigma\bigl(\mathbf{W}_{\!g}\,\mathbf{x}_{\text{rad},t} + b_g\bigr).
\label{eq:horizon}
\end{equation}
Here $\sigma$ is the logistic sigmoid, the scalar gate $\alpha_t$ scales every channel, $\mathbf{W}_{\!\Delta}^{\text{rad}}$ has the output shape of $\mathbf{W}_{\!\Delta}^{\text{img}}$, and $\mathbf{W}_{\!g} \in \mathbb{R}^{1 \times D}$ and $b_g \in \mathbb{R}$ form the gate. We initialise $\mathbf{W}_{\!\Delta}^{\text{rad}}$ and $\mathbf{W}_{\!g}$ at zero and set $b_g = -2$, so the gate opens at $\sigma(-2) \approx 0.12$ and the radar term starts at zero. A small step size makes $\bar{\mathbf{A}}_t$ close to the identity and retains the state over long ranges. A large step size resets the state and focuses on the current token~\cite{mamba}. The horizon pathway lets radar choose this balance locally. The small initial gate keeps the modulation latent at the start, so training opens it only if it helps.

\subsubsection{Readout modulation}
\label{sec:readout}

The readout pathway adds a radar term to $\mathbf{C}_t$,
\begin{equation}
\mathbf{C}_t = \mathbf{W}_{\!C}^{\text{img}}\,\mathbf{x}_{\text{img},t} + \mathbf{W}_{\!C}^{\text{rad}}\,\mathbf{x}_{\text{rad},t},
\label{eq:readout}
\end{equation}
where $\mathbf{W}_{\!C}^{\text{rad}}$ is initialised at zero. Horizon modulation changes what the state stores. Readout modulation changes what each token reports through $y_t = \mathbf{C}_t^{\!\top}\mathbf{h}_t$ and leaves the stored state untouched. No gate is needed because the projection grows only under gradient.

\subsubsection{Joint modulation}
\label{sec:joint}

The full RMS block applies Eq.~\eqref{eq:horizon} and Eq.~\eqref{eq:readout} in one Mamba layer. At initialisation $\mathbf{W}_{\!\Delta}^{\text{rad}} = \mathbf{W}_{\!C}^{\text{rad}} = \mathbf{0}$, so
\begin{equation}
\Dl_t\big|_{\text{init}} = \mathrm{softplus}\bigl(\mathbf{W}_{\!\Delta}^{\text{img}}\,\mathbf{x}_{\text{img},t}\bigr), \qquad \mathbf{C}_t\big|_{\text{init}} = \mathbf{W}_{\!C}^{\text{img}}\,\mathbf{x}_{\text{img},t}.
\label{eq:init-parity}
\end{equation}
This reproduces the image selection of Eq.~\eqref{eq:sel-img} exactly. The recurrence equals baseline Mamba at step zero, which we verify numerically.

\subsection{Multi-View Scan Pyramid (MVSP)}
\label{sec:mvsp}

Applying RMS at every decoder level is inefficient. Scan cost grows linearly with the number of tokens and exceeds $10^5$ tokens per direction at the finest levels. The spatial support of radar shrinks with scale, so the heaviest compute would land where radar helps least. MVSP matches the operator at each level to the reach of radar evidence. At the two coarsest levels, RMS runs over the full image in four directions and gathers context across the scene. This approximates a 2D receptive field at linear cost. At the middle level, RMS runs only in windows centred on projected radar returns. At the two finest levels, FiLM modulates features with radar at constant cost per pixel. Every level also ends with a radar FiLM block initialised at identity, which is the only fusion operator at the two finest levels. Scan compute goes where radar evidence reaches furthest.

\subsection{Training objective}
\label{sec:loss}

The objective combines robust supervision from a dense but noisy target with accurate supervision from a clean sparse one. Let $\mathbf{D}_{\text{pred}}$, $\mathbf{D}_{\text{main}}$, and $\mathbf{D}_{\text{sparse}}$ be the prediction, the main target, and an optional clean sparse target. On nuScenes, $\mathbf{D}_{\text{main}} = \mathbf{D}_{\text{acc}}$ is accumulated over 160 LiDAR sweeps and is dense but noisy, and $\mathbf{D}_{\text{sparse}} = \mathbf{D}_{\text{gt}}$ is a single clean sweep. On ZJU-4DRadarCam, $\mathbf{D}_{\text{main}}$ is the interpolated dense depth released with the dataset, and no $\mathbf{D}_{\text{sparse}}$ exists. Depths are normalised in log space to $\tilde{d} \in [-1,1]$ over $[d_{\min}, d_{\max}]$, and each loss uses only the valid pixels of its target. The objective is
\begin{equation}
\mathcal{L} = \lambda_{\log}\mathcal{L}_{\log} + \lambda_{\text{lin}}\mathcal{L}_{\text{lin}} + \lambda_{\text{grad}}\mathcal{L}_{\text{grad}} + \lambda_{\text{sparse}}\mathcal{L}_{\text{sparse}}.
\label{eq:loss}
\end{equation}
$\mathcal{L}_{\log}$ is an L1 loss in log space against $\mathbf{D}_{\text{sparse}}$, or against $\mathbf{D}_{\text{main}}$ when no sparse target exists. $\mathcal{L}_{\text{lin}}$ is a Huber loss ($\delta = 5$\,m) against $\mathbf{D}_{\text{main}}$. $\mathcal{L}_{\text{grad}}$ is an L1 loss on the $x$ and $y$ gradients of $\tilde{\mathbf{D}}_{\text{pred}} - \tilde{\mathbf{D}}_{\text{main}}$. $\mathcal{L}_{\text{sparse}}$ is an L1 loss in metres against $\mathbf{D}_{\text{sparse}}$, as in the sparse term of TacoDepth~\cite{TacoDepth}. Both linear losses are divided by $d_{\max}$. Log space is used only on the clean target. Beyond 5\,m the Huber loss grows linearly, which bounds the gradient of large residuals so that rare large residuals do not dominate training. We set $\lambda_{\log}=\lambda_{\text{lin}}=\lambda_{\text{sparse}}=1$ and $\lambda_{\text{grad}}=0.5$ unless stated otherwise.

\section{Experiments}
\label{sec:experiments}

We conduct experiments on the nuScenes~\cite{nuscenes} and ZJU-4DRadarCam~\cite{RadarCam-Depth} datasets, following prior radar camera depth work~\cite{RadarCam-Depth,TacoDepth,CaFNet,LiSBD,Singh,Lin,RC-PDA,R4Dyn,DORN}.

\subsection{Datasets and evaluation protocols}
\label{sec:datasets}

The nuScenes dataset~\cite{nuscenes} has 1{,}000 driving scenes with about 40{,}000 synchronised keyframes from Boston and Singapore, recorded with a vehicle platform carrying a camera, 3D radar, LiDAR, and an inertial measurement unit. We use the standard split of 700, 150, and 150 scenes for training, validation, and test from prior work~\cite{RadarCam-Depth,TacoDepth,CaFNet,LiSBD,Singh}. Front camera images are $900 \times 1600$, and each radar frame has 40 to 100 returns in the camera FoV.

ZJU-4DRadarCam~\cite{RadarCam-Depth} has 33{,}409 keyframes from a ground robot with a camera, a 4D radar, and a denser LiDAR. We use the official split of 29{,}312 frames for training and validation and 4{,}097 for test. As in prior work~\cite{RadarCam-Depth,TacoDepth}, we crop images from $720 \times 1280$ to $300 \times 1280$ to remove the hood and sky.

Following prior work, we evaluate against sparse LiDAR from a single sweep within 50, 70, and 80\,m~\cite{RadarCam-Depth,TacoDepth,CaFNet,LiSBD}. MAE and RMSE are the primary metrics. On ZJU and in the ablations we also report the inverse depth metrics iMAE and iRMSE, computed on $1/\hat{D}$ and $1/D_{\text{gt}}$ in km$^{-1}$. These metrics complement MAE and RMSE by weighting errors at short range more heavily.

\subsection{Implementation details}
\label{sec:impl}

On nuScenes, $\mathbf{D}_{\text{acc}}$ accumulates $\mathbf{D}_{\text{gt}}$ over 160 nearby LiDAR sweeps following~\cite{TacoDepth,LiSBD}. A light filter on the upper image removes accumulation artefacts from this target, and model inputs are unchanged. On ZJU-4DRadarCam, $\mathbf{D}_{\text{acc}}$ is the interpolated dense depth from~\cite{RadarCam-Depth} without filtering. The radar GSE uses three PCA-GM layers with 64 hidden channels. Each of the two coarsest MVSP levels stacks two RMS layers. The middle level uses windows of size $w = 8$ with Gaussian scatter weights of standard deviation $w/2.5$. We train with Adam at a learning rate of $10^{-4}$ and the step schedule of TacoDepth~\cite{TacoDepth}, which lowers the rate by $10^{-5}$ every ten epochs down to half its initial value. The batch size is 12 with bf16 autocast. Each run trains for at most 50 epochs. We evaluate every 2{,}000 steps, keep the checkpoint with the lowest MAE@$80$, and end a run once this metric has saturated. The depth range $[d_{\min}, d_{\max}]$ is $[0.5, 120]$\,m on nuScenes and $[0.5, 80]$\,m on ZJU-4DRadarCam. The ceiling of 120\,m keeps the roughly $15\%$ of nuScenes radar returns between 80 and 120\,m away from an artificial boundary. Evaluation stops at 80\,m on both datasets. Table~\ref{tab:main} reports the mean and standard deviation over three training seeds. Other experiments use one seed unless stated otherwise.

\subsection{Comparisons with state of the art methods}
\label{sec:sota}

Table~\ref{tab:main} compares SemoDepth with eight published methods on nuScenes under the protocol of TacoDepth~\cite{TacoDepth}, with the same split, $900{\times}1600$ input, LiDAR ground truth, and evaluation ranges. Published results are taken from the original papers. Our evaluation script reproduces the published CaFNet result from its public weights ($2116$ against $2109.8$\,mm MAE@$80$). We also reimplement FusionMamba~\cite{fusionmamba} (marked $\dagger$) as a modern Mamba fusion baseline. It was proposed for multimodal image fusion and has no radar camera depth result. We adapt its dual encoder backbone (VSSM\_Fusion) with cross modal SS2D fusion at every scale. We train it with the supervision, optimiser, loss, and schedule of SemoDepth at a matched budget of about 38\,M parameters. Latency is measured for every method with a runnable release on identical hardware (NVIDIA RTX PRO 6000 Blackwell, batch 1, FP32, $900{\times}1600$ input, mean over 100 iterations after 10 warmup iterations).

\begin{table}[t]
\centering
\small
\setlength{\tabcolsep}{4pt}
\caption{Comparison with published radar camera depth methods on nuScenes~\cite{nuscenes}. Time is single frame latency, measured on identical hardware for methods with a runnable release. Baseline MAE and RMSE come from the original publications. $^\dagger$Our reimplementation of FusionMamba for depth completion, trained with the supervision, optimiser, and loss of SemoDepth at a matched parameter budget. SemoDepth reports the mean and standard deviation over three seeds. Best per column in bold.}
\label{tab:main}
\resizebox{\linewidth}{!}{%
\begin{tabular}{@{}lcccccccc@{}}
\toprule
\multicolumn{1}{c}{\multirow{2}{*}{Method}} & \multirow{2}{*}{Time (ms)$\downarrow$} & \multicolumn{2}{c}{0--50\,m} & \multicolumn{2}{c}{0--70\,m} & \multicolumn{2}{c}{0--80\,m} \\
\cmidrule(lr){3-4} \cmidrule(lr){5-6} \cmidrule(lr){7-8}
 & & MAE (mm) $\downarrow$ & RMSE (mm) $\downarrow$ & MAE (mm) $\downarrow$ & RMSE (mm) $\downarrow$ & MAE (mm) $\downarrow$ & RMSE (mm) $\downarrow$ \\
\midrule
Lin et al.~\cite{Lin} (IROS'20)        &    -- & 2034.9 & 4316.5 & 2294.7 & 5338.2 & 2371.0 & 5623.0 \\
RC-PDA~\cite{RC-PDA} (CVPR'21)         & 145.0 & 2225.0 & 4156.5 & 3326.1 & 6700.6 & 3713.6 & 7692.8 \\
R4Dyn~\cite{R4Dyn} (3DV'21)            &    -- & -- & -- & -- & -- & -- & 6434.0 \\
DORN~\cite{DORN} (ICIP'21)             &  60.9 & 1926.6 & 4124.8 & 2380.6 & 5252.7 & 2467.7 & 5554.3 \\
Singh et al.~\cite{Singh} (CVPR'23)    &  40.9 & 1727.7 & 3746.8 & 2073.2 & 4590.7 & 2179.3 & 4898.7 \\
CaFNet~\cite{CaFNet} (IROS'24)         &  44.2 & 1674.2 & 3674.5 & 2010.3 & 4493.1 & 2109.8 & 4765.6 \\
Li et al.~\cite{LiSBD} (ECCV'24)       &  46.4 & 1524.5 & 3567.3 & 1822.9 & 4303.6 & 1927.0 & 4609.6 \\
FusionMamba$^\dagger$~\cite{fusionmamba} (VI'24) & -- & 1179.9 & 3337.2 & 1483.3 & 4282.5 & 1590.5 & 4640.0 \\
TacoDepth~\cite{TacoDepth} (CVPR'25)   &    -- & 1423.6 & 3275.8 & 1712.6 & 3960.5 & 1833.4 & 4150.2 \\
\midrule
SemoDepth (Ours)                       & \textbf{26.8} & $\mathbf{940.1}_{\pm 4.1}$ & $\mathbf{2785.7}_{\pm 2.5}$ & $\mathbf{1199.8}_{\pm 1.5}$ & $\mathbf{3622.4}_{\pm 16.5}$ & $\mathbf{1285.2}_{\pm 0.9}$ & $\mathbf{3935.4}_{\pm 21.5}$ \\
\bottomrule
\end{tabular}}
\end{table}

SemoDepth has the lowest MAE and RMSE at every range. Across the published methods in Table~\ref{tab:main}, MAE@$80$ fell by about $540$\,mm in five years, from $2371$\,mm (Lin et al.) to $1833$\,mm (TacoDepth). SemoDepth lowers it by another $548$\,mm in one step. The gain does not come from our supervision configuration. Trained on the raw target accumulated over 160 sweeps without the gradient term, SemoDepth reaches MAE@$80$ of $1311.9$\,mm, $28.4\%$ below TacoDepth. \FM{} reaches $1587.8$\,mm with the raw configuration and $1590.5$\,mm with the cleaned one. The gap between the two architectures is $17.4\%$ with raw and $19.2\%$ with cleaned supervision.

Two patterns in Table~\ref{tab:main} tie the result to the design claims of \S\ref{sec:method}. MAE gains are more than twice the RMSE gains at every range ($34.0\%$ against $15.0\%$ within 50\,m, $29.9\%$ against $8.5\%$ within 70\,m, and $29.9\%$ against $5.2\%$ within 80\,m). MAE reflects the typical pixel, while RMSE stresses rare large errors, so the gain accrues mostly to the typical pixel rather than the worst case. This is the behaviour expected from selection inside the scan. By reshaping $\Dl$ and $\mathbf{C}$ at every token, RMS lets the metric anchor of each radar return propagate along the whole sequence rather than only near its projected pixel. Such reshaping at the level of tokens is hard to achieve with fusion outside the scan.

\begin{figure}[t]
\centering
\includegraphics[width=\linewidth]{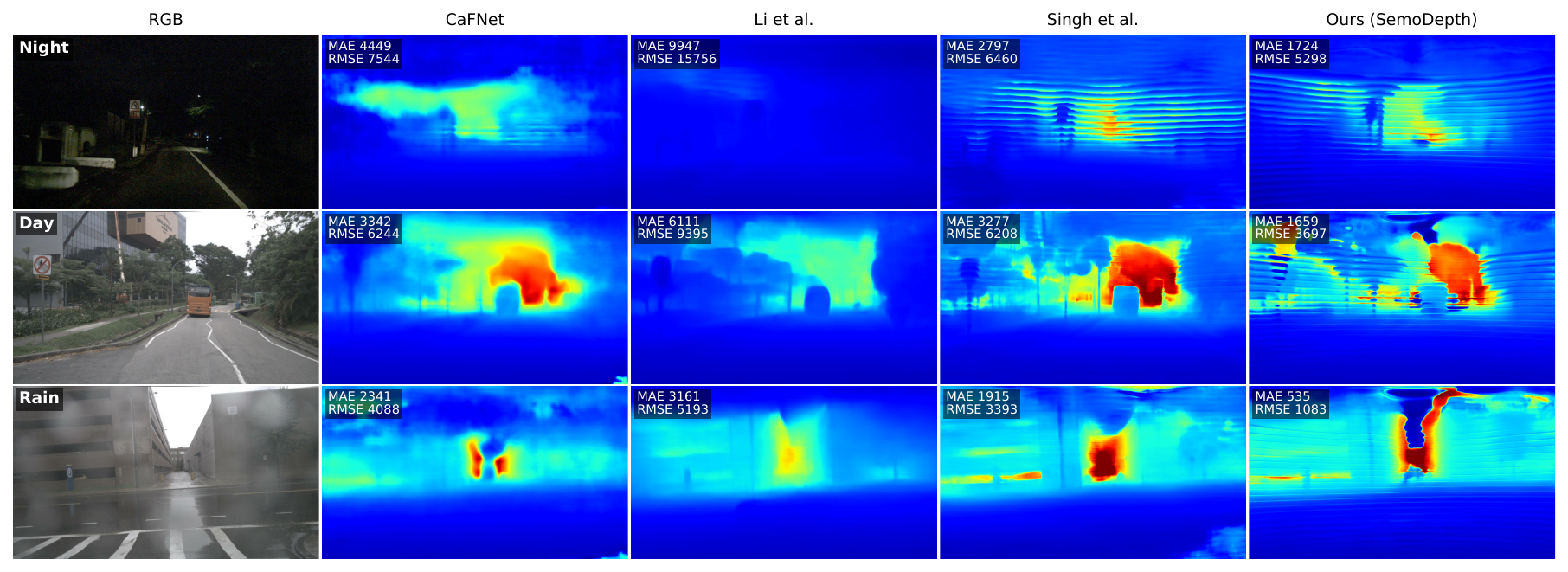}
\caption{Qualitative comparison on nuScenes~\cite{nuscenes}. Rows show night (frame 5319), day (frame 161), and rain (frame 3088). Columns show the RGB input, CaFNet~\cite{CaFNet}, Li et al.~\cite{LiSBD}, Singh et al.~\cite{Singh}, and SemoDepth. Each panel gives the MAE and RMSE of the frame in millimetres against sparse LiDAR within 80\,m, and the colour map is clipped at 80\,m. SemoDepth has the lowest MAE and RMSE on every frame. The largest margin is in rain, with $535$\,mm MAE against $1915$\,mm for the next best baseline.}
\label{fig:qualitative}
\end{figure}

The second pattern is the gap to TacoDepth, the most informative comparison across methods. The windowed flash attention of TacoDepth around radar points is architecturally closest to the windowed RMS of our middle tier. A steady MAE margin of about $30\%$ at every range is unlikely to come from windowed processing alone. It is consistent with the two tiers that distinguish MVSP. The first is the scan across the whole scene at the coarsest levels, which has no analogue in the pyramid of TacoDepth. The second is the FiLM at constant cost that keeps radar active at the finest levels, where the pyramid of TacoDepth attenuates it. \FM{} gives a second informative comparison. Trained with matched supervision, optimiser, and loss, it reaches MAE@$80$ of $1591$\,mm, between TacoDepth and SemoDepth. A Mamba backbone alone closes part of the gap, but a margin of $19.2\%$ MAE@$80$ remains. With the backbone family fixed and the training pipeline matched, this margin isolates the contribution of selection inside the scan over fusion before it. It is the distinction drawn in \S\ref{sec:related-ssm}, now visible as a measured gap rather than a design argument. Fig.~\ref{fig:qualitative} shows examples at night, by day, and in rain against the three baselines with public code (CaFNet~\cite{CaFNet}, Li et al.~\cite{LiSBD}, and Singh et al.~\cite{Singh}).

\begin{table}[t]
\centering
\small
\setlength{\tabcolsep}{4pt}
\caption{Comparison on ZJU-4DRadarCam~\cite{RadarCam-Depth}. Baseline numbers are as reported by TacoDepth~\cite{TacoDepth}. No method here uses a monocular prior. Best per column in bold.}
\label{tab:zju}
\resizebox{\linewidth}{!}{%
\begin{tabular}{@{}lcccccc@{}}
\toprule
\multicolumn{1}{c}{\multirow{2}{*}{Method}} & \multicolumn{2}{c}{0--50\,m} & \multicolumn{2}{c}{0--70\,m} & \multicolumn{2}{c}{0--80\,m} \\
\cmidrule(lr){2-3} \cmidrule(lr){4-5} \cmidrule(lr){6-7}
 & MAE (mm) $\downarrow$ & RMSE (mm) $\downarrow$ & MAE (mm) $\downarrow$ & RMSE (mm) $\downarrow$ & MAE (mm) $\downarrow$ & RMSE (mm) $\downarrow$ \\
\midrule
DORN~\cite{DORN}                     & 2210.2          & 4129.7          & 2402.2          & 4625.2          & 2447.6          & 4760.0          \\
Singh et al.~\cite{Singh}            & 1785.4          & 3704.6          & 1932.7          & 4137.1          & 1979.5          & 4309.3          \\
TacoDepth~\cite{TacoDepth}           & 1120.1          & 2686.7          & 1181.8          & \textbf{2906.3} & 1201.1          & \textbf{2990.7} \\
SemoDepth (Ours)                     & \textbf{1029.2} & \textbf{2631.6} & \textbf{1111.6} & 2946.9          & \textbf{1137.2} & 3053.0          \\
\bottomrule
\end{tabular}}
\end{table}

On ZJU-4DRadarCam (Table~\ref{tab:zju}), every method uses one radar scan and one image per frame, the same released training target, and the same sparse evaluation. SemoDepth has the best MAE at every range and the best RMSE within 50\,m. TacoDepth keeps a small edge on RMSE within 70\,m ($1.4\%$) and 80\,m ($2.1\%$). Among methods that also consume a monocular depth prior, the best of five RadarCam-Depth configurations~\cite{RadarCam-Depth} reaches $1183.5$\,mm MAE@$80$ and $3229.0$\,mm RMSE@$80$, and SemoDepth beats both without such a prior. TacoDepth with a monocular prior remains stronger at $1032.5$\,mm MAE@$80$. The RMSE gap to TacoDepth comes from a small far tail. Pixels beyond 70\,m make up $0.4\%$ of our test pixels but $7.5\%$ of the squared error, and SemoDepth under predicts them by $7.3$\,m on average between 70 and 80\,m. For the $78.6\%$ of pixels within 20\,m, the median error is below $0.3$\,m. On inverse depth, which weights the far tail less, SemoDepth leads TacoDepth by $18\%$ on iMAE and $8\%$ on iRMSE.

\begin{table}[t]
\centering
\footnotesize
\setlength{\tabcolsep}{6pt}
\caption{Inference efficiency on nuScenes~\cite{nuscenes}. Time follows the protocol of Table~\ref{tab:main}. TacoDepth has no public release. Params and FLOPs come from the TacoDepth paper for DORN, Singh et al., and TacoDepth. We measure them with thop ($2\times$MACs) for CaFNet and Li et al. The FLOPs of SemoDepth are a lower bound from fvcore. Best per column in bold.}
\label{tab:efficiency}
\begin{tabular}{@{}lccc@{}}
\toprule
Method & Params (M) $\downarrow$ & FLOPs (G) $\downarrow$ & Time (ms) $\downarrow$ \\
\midrule
DORN~\cite{DORN} (ICIP'21)           & 183.4         & 923.5         & 60.9 \\
Singh et al.~\cite{Singh} (CVPR'23)  & 22.8          & 502.1         & 40.9 \\
CaFNet~\cite{CaFNet} (IROS'24)       & 62.2          & 1426.6        & 44.2 \\
Li et al.~\cite{LiSBD} (ECCV'24)     & 36.9          & \textbf{31.0} & 46.4 \\
TacoDepth~\cite{TacoDepth} (CVPR'25) & \textbf{13.5} & 139.3         & --   \\
SemoDepth (Ours)                     & 38.1          & 271.8         & \textbf{26.8} \\
\bottomrule
\end{tabular}
\end{table}

\subsection{Inference efficiency}
\label{sec:efficiency}

SemoDepth has the lowest latency among methods with a public release ($26.8$\,ms), 1.5 to 2.3 times faster than Singh et al., CaFNet, Li et al., and DORN (Table~\ref{tab:efficiency}). It has more parameters than TacoDepth and sits in the middle on FLOPs. MVSP keeps scan compute at the coarse levels.

\subsection{Ablation studies}
\label{sec:ablation}

Table~\ref{tab:ablation} varies the decoder and the site where radar enters the scan. All rows share the ResNet-34 encoder, the radar GSE, the training recipe of \S\ref{sec:impl}, and the depth range.

\begin{table}[t]
\centering
\small
\setlength{\tabcolsep}{3.5pt}
\caption{Ablation on nuScenes~\cite{nuscenes}. The upper block changes the decoder. Uniform FiLM replaces the scans of Tiers 2 and 3 with FiLM at every level. Full RMS to level 2 replaces the windowed scan of Tier 2 with the full image scan. The lower block changes which selection parameters radar modulates inside the MVSP decoder. In the row without radar in the scan, radar still reaches the decoder through FiLM. The blend row adds a confidence gated blend in the style of FusionMamba~\cite{fusionmamba,mambadfuse} before the scan. Best per column in bold.}
\label{tab:ablation}
\resizebox{\linewidth}{!}{%
\begin{tabular}{@{}lcccccc@{}}
\toprule
 & \multicolumn{2}{c}{0--50\,m} & \multicolumn{2}{c}{0--70\,m} & \multicolumn{2}{c}{0--80\,m} \\
\cmidrule(lr){2-3} \cmidrule(lr){4-5} \cmidrule(lr){6-7}
Variant & MAE (mm) $\downarrow$ & RMSE (mm) $\downarrow$ & MAE (mm) $\downarrow$ & RMSE (mm) $\downarrow$ & MAE (mm) $\downarrow$ & RMSE (mm) $\downarrow$ \\
\midrule
Uniform FiLM                                   & 1094         & 3075          & 1437          & 4166          & 1564          & 4668          \\
Full RMS to level 2                            & \textbf{919} & \textbf{2728} & 1206          & 3649          & 1304          & 4011          \\
\midrule
No radar in the scan                           & 950          & 2787          & 1217          & 3674          & 1306          & 4004          \\
$\mathbf{B}$ only                              & 933          & 2782          & 1217          & 3677          & 1315          & 4025          \\
$\Dl$ only (horizon)                           & 964          & 2755          & 1251          & 3614          & 1352          & 3953          \\
$\mathbf{C}$ only (readout)                    & 956          & 2792          & 1227          & 3672          & 1317          & 4012          \\
$\Dl{+}\mathbf{B}{+}\mathbf{C}$                & 967          & 2802          & 1267          & 3748          & 1360          & 4092          \\
$\Dl{+}\mathbf{C}$ with blend before the scan  & 940          & 2749          & 1211          & 3657          & 1304          & 4019          \\
$\Dl{+}\mathbf{C}$ (joint, main)               & 943          & 2784          & \textbf{1201} & \textbf{3611} & \textbf{1286} & \textbf{3920} \\
\bottomrule
\end{tabular}}
\end{table}

\paragraph{MVSP is critical for accuracy, not only for speed.} The Uniform FiLM row replaces the scans of Tiers 2 and 3 with FiLM at every level and keeps every other component fixed. MVSP lowers MAE@$80$ from $1564$ to $1286$\,mm ($18\%$), so the pyramid plays a central role in accuracy. We also extend the full image scan with four directions down the decoder (Table~\ref{tab:fullrms}). At level 2 it replaces the windowed scan, and at levels 1 and 0 it replaces FiLM. Each finer level quadruples the number of tokens. Extending the scan to level 2 gains $2.6\%$ MAE within 50\,m. It loses $1.4\%$ MAE@$80$ and $2.3\%$ RMSE@$80$ and adds $27\%$ latency and $4.6$\,M parameters, so it is worse on the headline metrics. The scan at fine resolution sharpens nearby structure but dilutes the advantage at long range. Levels 1 and 0 multiply latency by $3.1$ and $10.5$, and level 0 needs $5.4$ times the memory. The MVSP allocation is the better balance of accuracy and cost.

\begin{table}[t]
\centering
\footnotesize
\setlength{\tabcolsep}{5pt}
\caption{Full RMS at finer decoder levels on nuScenes~\cite{nuscenes}. Latency is relative to MVSP, measured at batch 1 and FP32. Levels 1 and 0 are compared on cost only.}
\label{tab:fullrms}
\begin{tabular}{@{}lccccc@{}}
\toprule
Full RMS down to level & Latency & Peak memory (GB) & Params (M) & MAE@$80$ (mm) & RMSE@$80$ (mm) \\
\midrule
3 (MVSP, main)  & $1.00\times$  & 2.16  & 38.1 & 1285.8 & 3920.2 \\
2               & $1.27\times$  & 2.17  & 42.7 & 1303.8 & 4011.4 \\
1               & $3.15\times$  & 3.59  & 47.9 & -- & -- \\
0               & $10.53\times$ & 11.66 & 53.1 & -- & -- \\
\bottomrule
\end{tabular}
\end{table}

\paragraph{Both RMS pathways contribute.} Within MVSP, horizon modulation and readout modulation each recover much of the gap between Uniform FiLM and the joint model, so the pathways are not redundant. Their behaviour is complementary. Horizon modulation (\S\ref{sec:horizon}) controls how information flows through the recurrence. Readout modulation (\S\ref{sec:readout}) controls what is extracted at the emission step. Their combination is needed for full cross modal expressivity. Joint modulation attains the lowest MAE@$80$ ($1286$\,mm against $1352$ for horizon only and $1317$ for readout only) and the lowest RMSE@$80$ ($3920$\,mm against $3953$ and $4012$). On RMSE within 50\,m, horizon modulation alone edges out the joint model by about $1\%$, which we attribute to minor optimisation variance.

\paragraph{The input projection should stay driven by the image.} With radar only in $\mathbf{B}$, MAE@$80$ is $1315$\,mm, within 9\,mm of the variant without radar in the scan ($1306$\,mm). Adding $\mathbf{B}$ to $\Dl$ and $\mathbf{C}$ raises MAE@$80$ by $5.8\%$ to $1360$\,mm, worse than no radar in the scan. Radar in $\mathbf{B}$ alone does lower MAE within 50\,m, but it does not carry to longer ranges. These results support keeping the input projection driven by the image.

\paragraph{Fusion before the scan adds nothing on top of RMS.} The blend row adds a FusionMamba style blend before the scan to the joint model. It reaches $1304$\,mm MAE@$80$ against $1286$\,mm for the joint model and leads only within 50\,m. Over three seeds, the blend gives $1305.3 \pm 11.5$\,mm MAE@$80$ and the joint model gives $1285.2 \pm 0.9$\,mm. The blend is worse on average and an order of magnitude less stable across seeds. Once radar modulates selection inside the scan, an extra pathway before the scan adds no signal at the ranges that drive the headline metrics. This complements the comparison with the full FusionMamba backbone in Table~\ref{tab:main}.

\subsection{Robustness to radar corruption and conditions}
\label{sec:robustness}

\begin{figure}[t]
\centering
\includegraphics[width=\linewidth]{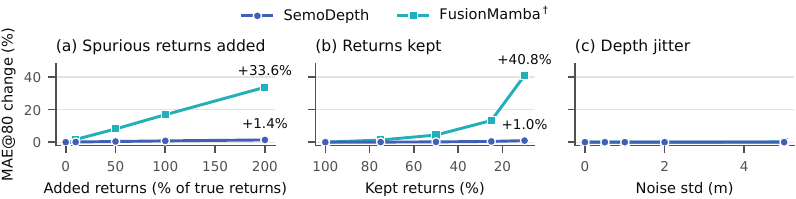}
\caption{Change in MAE@$80$ on nuScenes under radar corruption at test time. (a) Spurious returns added, as a share of the true returns. (b) Share of true returns kept. (c) Gaussian noise on the depth of true returns. Both models see identical corrupted inputs. Removing radar entirely costs SemoDepth $3.3\%$ and \FM{} $155\%$.}
\label{fig:robustness}
\end{figure}

We test whether selection filters out corrupted radar. A deterministic procedure corrupts the radar of each nuScenes validation frame with a seed tied to the frame index, so every model sees identical inputs. We compare SemoDepth with \FM{}, which shares supervision, loss, recipe, and parameter budget but fuses radar before the scan. Spurious returns are placed at random image positions with random depths between 0.5 and 120\,m. Removal keeps a random subset of true returns. Depth noise perturbs the true returns with Gaussian noise (Fig.~\ref{fig:robustness}). With twice as many spurious returns as true ones, SemoDepth loses $1.4\%$ MAE@$80$ and \FM{} loses $33.6\%$. Keeping $10\%$ of the returns costs $1.0\%$ and $40.8\%$. Noise with a standard deviation of 5\,m costs at most $0.2\%$ for either model, so both depend on where returns are rather than on their exact depth. With radar removed, SemoDepth loses $3.3\%$ and reaches $1327$\,mm, still below every published method in Table~\ref{tab:main}. Inside the scans, radar acts only through modulations of $\Dl$ and $\mathbf{C}$ that start at zero and never enters the token stream or $\mathbf{B}$. Elsewhere it enters through FiLM blocks that start at identity. These narrow, learned channels are consistent with the small effect of corrupted radar.

\begin{table}[t]
\centering
\footnotesize
\setlength{\tabcolsep}{5pt}
\caption{nuScenes validation frames split by condition, within 80\,m. A scene counts as night or rain when its description contains that word. Rain overlaps with day and night. Both models are those evaluated in Fig.~\ref{fig:robustness}. Best per row and metric in bold.}
\label{tab:conditions}
\begin{tabular}{@{}lcccccc@{}}
\toprule
 & \multicolumn{3}{c}{SemoDepth} & \multicolumn{3}{c}{\FM{}} \\
\cmidrule(lr){2-4} \cmidrule(lr){5-7}
Split (frames) & MAE (mm) & RMSE (mm) & iMAE & MAE (mm) & RMSE (mm) & iMAE \\
\midrule
Day (5{,}282)   & \textbf{1177.0} & \textbf{3731.1} & \textbf{3.96} & 1530.0          & 4532.8          & 5.06 \\
Night (587)     & 2252.0          & 5925.5          & \textbf{7.41} & \textbf{2135.2} & \textbf{5605.2} & 7.83 \\
Rain (1{,}061)  & \textbf{1681.7} & \textbf{4834.3} & \textbf{4.78} & 1980.3          & 5409.6          & 5.85 \\
All (5{,}869)   & \textbf{1284.5} & \textbf{3950.6} & \textbf{4.30} & 1590.5          & 4640.0          & 5.34 \\
\bottomrule
\end{tabular}
\end{table}

Table~\ref{tab:conditions} splits the validation frames by condition. The night split follows TacoDepth~\cite{TacoDepth}. nuScenes contains no fog scenes, so fog cannot be evaluated. SemoDepth leads by $23\%$ MAE@$80$ by day and by $15\%$ in rain, where the camera degrades and radar does not. At night \FM{} is about $5\%$ better on MAE@$80$ and RMSE@$80$, while SemoDepth keeps the lower iMAE. The night split holds only 587 frames, and its result varies with the training seed. A second seed of the main configuration reaches $2146$\,mm MAE@$80$ at night.

\section{Limitations}
\label{sec:limitations}

Four limits qualify our claims. First, at night the camera signal is weak. On the 587 night frames, fusion before the scan can match or beat SemoDepth on linear metrics, and the outcome varies with the seed. Second, supervision on nuScenes comes from LiDAR accumulated over 160 sweeps, which leaves trails behind moving objects. Our filtered target, used with gradient matching, lowers RMSE@$80$ by 7 to $10\%$ for both architectures relative to the raw configuration, yet some artefacts remain visible in the predicted maps. Third, published methods are compared through their reported numbers. TacoDepth has no public release, so we control for training effects with \FM{}, retrained under matched settings with both the raw and the cleaned target. Fourth, SemoDepth under predicts far depth. On ZJU-4DRadarCam the mean bias between 70 and 80\,m is $-7.3$\,m, and this tail explains its RMSE gap at 70 and 80\,m. We have not tested other sensor pairs, fog, or embedded hardware. Dense depth errors can propagate to planning, so any deployment needs evaluation by scene type and monitoring of inputs that differ from the training data.

\section{Conclusion}

We argued that radar should enter a Mamba depth model through its selection mechanism rather than its input stream. Radar-Modulated Selection and the Multi-View Scan Pyramid realise this idea. SemoDepth sets a new state of the art on nuScenes for MAE and RMSE at every range and has the lowest latency among methods with a public release. Controlled studies show that modulating $\Dl$ and $\mathbf{C}$ helps, modulating $\mathbf{B}$ does not improve MAE@$80$, and a blend before the scan adds nothing. Under spurious, sparse, or missing radar, SemoDepth degrades far less than a Mamba model that fuses radar before the scan. In sequence models whose recurrence depends on the input, the selection mechanism itself can serve as the interface between modalities. We expect this idea to extend to other pairs of sparse and dense sensors, such as LiDAR and camera or radar and thermal imaging, and to tasks beyond depth. Our findings suggest that cross modal learning in sequence models should be reconsidered from the perspective of selection rather than representation. This opens a new direction for multimodal sequence modelling beyond vision.

\begin{ack}
This work was supported by Japan Science and Technology Agency (JST) Adopting Sustainable Partnerships for Innovative Research Ecosystem (ASPIRE) Grant Number JPMJAP2326, Japan. The authors declare no competing interests.
\end{ack}

\bibliographystyle{plainnat}
\bibliography{references}

\appendix

\section{Notation}
\label{app:notation}

Table~\ref{tab:notation} collects the symbols used in the paper.

\begin{table}[ht]
\centering
\small
\caption{Symbols used in the paper.}
\label{tab:notation}
\begin{tabular}{@{}ll@{}}
\toprule
Symbol & Meaning \\
\midrule
$\mathbf{x}_{\text{img},t},\ \mathbf{x}_{\text{rad},t} \in \mathbb{R}^{D}$ & Image and radar tokens at position $t$ of a scanned sequence \\
$L,\ D,\ N$ & Sequence length, number of channels, state dimension \\
$x_t,\ y_t$ & Scalar input and output of one channel of the scan \\
$\mathbf{h}_t \in \mathbb{R}^{N}$ & Hidden state of one channel \\
$\mathbf{A} \in \mathbb{R}^{D \times N}$ & State matrix, trained and shared by all tokens \\
$\bar{\mathbf{A}}_t,\ \bar{\mathbf{B}}_t$ & State and input matrices after ZOH discretization \\
$\Dl_t \in \mathbb{R}^{D}$ & Step size, which sets the memory horizon \\
$\mathbf{B}_t \in \mathbb{R}^{N}$ & Input projection, computed from the image only \\
$\mathbf{C}_t \in \mathbb{R}^{N}$ & Readout of the hidden state \\
$\mathbf{W}_{\!\Delta}^{\text{img}},\ \mathbf{W}_{\!B},\ \mathbf{W}_{\!C}^{\text{img}}$ & Image projections that produce $\Dl_t$, $\mathbf{B}_t$, $\mathbf{C}_t$ \\
$\mathbf{W}_{\!\Delta}^{\text{rad}},\ \mathbf{W}_{\!C}^{\text{rad}}$ & Radar projections, initialised at zero \\
$\alpha_t,\ \mathbf{W}_{\!g},\ b_g$ & Radar gate on the step size, its weight and bias ($b_g = -2$) \\
$\sigma(\cdot)$ & Logistic sigmoid \\
$c_i,\ r_i$ & Image and radar features at decoder level $i$ \\
$H,\ W,\ d$ & Image height and width, range of a radar return \\
$d_{\min},\ d_{\max}$ & Depth range of the model \\
$w$ & Window size of the windowed scan at the middle level \\
$\mathbf{D}_{\text{pred}},\ \mathbf{D}_{\text{main}},\ \mathbf{D}_{\text{sparse}}$ & Prediction, main target, clean sparse target \\
$\mathbf{D}_{\text{acc}},\ \mathbf{D}_{\text{gt}}$ & LiDAR accumulated over 160 sweeps, LiDAR from a single sweep \\
$\tilde{\mathbf{D}}$ & Depth normalised in log space to $[-1, 1]$ \\
$\lambda_{\log},\ \lambda_{\text{lin}},\ \lambda_{\text{grad}},\ \lambda_{\text{sparse}}$ & Loss weights \\
\bottomrule
\end{tabular}
\end{table}

\section{Additional results on nuScenes}

\subsection{Day and night against published methods}
\label{app:daynight}

Table~\ref{tab:daynight} follows the day and night protocol of TacoDepth~\cite{TacoDepth}. A scene counts as night when its description contains ``night'', which splits the validation set into $5{,}282$ day and $587$ night keyframes. Numbers for Singh et al.\ and TacoDepth come from TacoDepth. SemoDepth is the main configuration of Table~\ref{tab:ablation}, evaluated against sparse LiDAR from a single sweep with predictions clamped to $[0.5, 80]$\,m.

\begin{table}[ht]
\centering
\small
\caption{Day and night results on nuScenes~\cite{nuscenes}. Following TacoDepth~\cite{TacoDepth}, scenes whose description contains ``night'' form the night split. Best per column within each split in bold.}
\label{tab:daynight}
\setlength{\tabcolsep}{3pt}
\resizebox{\linewidth}{!}{%
\begin{tabular}{@{}llrrrrrr@{}}
\toprule
\multirow{2}{*}{Scene} & \multirow{2}{*}{Method}
& \multicolumn{2}{c}{0--50\,m} & \multicolumn{2}{c}{0--70\,m}
& \multicolumn{2}{c}{0--80\,m} \\
\cmidrule(lr){3-4}\cmidrule(lr){5-6}\cmidrule(lr){7-8}
& & MAE (mm) $\downarrow$ & RMSE (mm) $\downarrow$ & MAE (mm) $\downarrow$ & RMSE (mm) $\downarrow$ & MAE (mm) $\downarrow$ & RMSE (mm) $\downarrow$ \\
\midrule
\multirow{3}{*}{Day}
& Singh et al.~\cite{Singh} (CVPR'23)  & 1618.9 & 3613.0 & 1924.7 & 4359.2 & 2017.9 & 4632.5 \\
& TacoDepth~\cite{TacoDepth} (CVPR'25) & 1389.5 & 3227.3 & 1680.9 & 3897.1 & 1782.4 & 4092.3 \\
& SemoDepth (Ours)                     & \textbf{857.9} & \textbf{2606.3} & \textbf{1105.2} & \textbf{3416.5} & \textbf{1190.2} & \textbf{3731.1} \\
\midrule
\multirow{3}{*}{Night}
& Singh et al.~\cite{Singh} (CVPR'23)  & 2340.8 & 4683.8 & 2863.9 & 5935.4 & 3012.9 & 6338.3 \\
& TacoDepth~\cite{TacoDepth} (CVPR'25) & \textbf{1673.6} & \textbf{3631.4} & \textbf{1944.8} & \textbf{4425.3} & 2207.6 & \textbf{4574.8} \\
& SemoDepth (Ours)                     & 1709.4 & 4381.5 & 2061.9 & 5358.2 & \textbf{2146.1} & 5621.7 \\
\midrule
\multirow{3}{*}{All}
& Singh et al.~\cite{Singh} (CVPR'23)  & 1727.7 & 3746.8 & 2073.2 & 4590.7 & 2179.3 & 4898.7 \\
& TacoDepth~\cite{TacoDepth} (CVPR'25) & 1423.6 & 3275.8 & 1712.6 & 3960.5 & 1833.4 & 4150.2 \\
& SemoDepth (Ours)                     & \textbf{943.0} & \textbf{2783.9} & \textbf{1200.9} & \textbf{3610.7} & \textbf{1285.8} & \textbf{3920.2} \\
\bottomrule
\end{tabular}}
\end{table}

By day (90\% of frames) SemoDepth wins every column. Its MAE is 33 to 38\% below TacoDepth and 41 to 47\% below Singh et al. Within 80\,m it lowers MAE by 33\% ($1190$ against $1782$\,mm) and RMSE by 9\% ($3731$ against $4092$\,mm) relative to TacoDepth. The day split dominates the aggregate numbers of Table~\ref{tab:main}.

At night (10\% of frames) SemoDepth leads Singh et al.\ in all six columns. Against TacoDepth it has the better MAE@$80$ ($2146$ against $2208$\,mm) but trails at shorter ranges and on RMSE. Another training seed, evaluated in Table~\ref{tab:conditions}, reaches $2252$\,mm MAE@$80$ at night, which trails TacoDepth. The ranking at night depends on the seed, and we do not claim a lead there. At night the camera signal is weakest, and large errors that dominate RMSE become more common. A strong attention based fusion competes closely in this regime.

\subsection{Complete ablation results}
\label{sec:ablation_full}

Table~\ref{tab:ablation_full} extends Table~\ref{tab:ablation} with inverse depth metrics and with the three seed results for the two configurations that carry the claim about fusion before the scan. MVSP lowers iMAE@$80$ from $5.14$ to $4.22$ ($18\%$) relative to Uniform FiLM, which matches the MAE@$80$ result. Radar only in $\mathbf{B}$ gives the lowest iMAE@$80$ and iRMSE@$80$. Inverse metrics weight near pixels most, and this variant is also strong within 50\,m. It does not help at 70 and 80\,m, where it falls behind the joint model and is close to the variant without radar in the scan. Capacity does not predict accuracy here. The variant with $\Dl$, $\mathbf{B}$, and $\mathbf{C}$ (38.2\,M parameters) and full RMS to level 2 (42.7\,M) both trail the main model (38.1\,M). We did not modulate $\mathbf{A}$ directly. It is a trained constant, and modulating $\Dl$ already changes the discretized transition $\bar{\mathbf{A}}_t = \exp(\Dl_t \mathbf{A})$ at every token.

\begin{table}[ht]
\centering
\small
\setlength{\tabcolsep}{3pt}
\footnotesize
\caption{Complete ablation on nuScenes~\cite{nuscenes}, including iMAE and iRMSE within 80\,m. Single seed rows match Table~\ref{tab:ablation}. The last two rows give the mean and standard deviation over three seeds. Best per column among single seed rows in bold.}
\label{tab:ablation_full}
\resizebox{\linewidth}{!}{%
\begin{tabular}{@{}lcccccccc@{}}
\toprule
 & \multicolumn{2}{c}{0--50\,m} & \multicolumn{2}{c}{0--70\,m} & \multicolumn{4}{c}{0--80\,m} \\
\cmidrule(lr){2-3} \cmidrule(lr){4-5} \cmidrule(lr){6-9}
Variant & MAE (mm) $\downarrow$ & RMSE (mm) $\downarrow$ & MAE (mm) $\downarrow$ & RMSE (mm) $\downarrow$ & MAE (mm) $\downarrow$ & RMSE (mm) $\downarrow$ & iMAE (km$^{-1}$) $\downarrow$ & iRMSE (km$^{-1}$) $\downarrow$ \\
\midrule
Uniform FiLM                                   & 1094         & 3075          & 1437          & 4166          & 1564          & 4668          & 5.14          & 13.44 \\
Full RMS to level 2                            & \textbf{919} & \textbf{2728} & 1206          & 3649          & 1304          & 4011          & 4.13          & 10.74 \\
No radar in the scan                           & 950          & 2787          & 1217          & 3674          & 1306          & 4004          & 4.25          & 11.01 \\
$\mathbf{B}$ only                              & 933          & 2782          & 1217          & 3677          & 1315          & 4025          & \textbf{3.98} & \textbf{10.37} \\
$\Dl$ only (horizon)                           & 964          & 2755          & 1251          & 3614          & 1352          & 3953          & 4.41          & 10.74 \\
$\mathbf{C}$ only (readout)                    & 956          & 2792          & 1227          & 3672          & 1317          & 4012          & 4.36          & 10.89 \\
$\Dl{+}\mathbf{B}{+}\mathbf{C}$                & 967          & 2802          & 1267          & 3748          & 1360          & 4092          & 4.22          & 10.74 \\
$\Dl{+}\mathbf{C}$ with blend before the scan  & 940          & 2749          & 1211          & 3657          & 1304          & 4019          & 4.29          & 11.03 \\
$\Dl{+}\mathbf{C}$ (joint, main)               & 943          & 2784          & \textbf{1201} & \textbf{3611} & \textbf{1286} & \textbf{3920} & 4.22          & 10.85 \\
\midrule
Blend before the scan, 3 seeds                 & $940.3_{\pm 10.3}$ & $2775.3_{\pm 23.6}$ & $1214.1_{\pm 10.9}$ & $3675.3_{\pm 19.1}$ & $1305.3_{\pm 11.5}$ & $4017.3_{\pm 6.3}$ & $4.29_{\pm 0.05}$ & $10.96_{\pm 0.11}$ \\
Joint (main), 3 seeds                          & $940.1_{\pm 4.1}$  & $2785.7_{\pm 2.5}$  & $1199.8_{\pm 1.5}$  & $3622.4_{\pm 16.5}$ & $1285.2_{\pm 0.9}$  & $3935.4_{\pm 21.5}$ & -- & -- \\
\bottomrule
\end{tabular}}
\end{table}

\subsection{Effect of the supervision configuration}
\label{app:cleaning}

The main SemoDepth result is trained on the filtered target accumulated over 160 sweeps with $\lambda_{\text{grad}}=0.5$. Table~\ref{tab:cleaning} compares this configuration with the raw target and $\lambda_{\text{grad}}=0$ for the three RMS variants and for \FM{}. The two configurations differ in both the target and the gradient term, so the table measures their combined effect. Gradient matching is used only with the cleaned target, because on the raw target it would reinforce the edges of accumulation trails. Each row uses the checkpoint selected by the rule of \S\ref{sec:impl}.

\begin{table}[ht]
\centering
\small
\caption{Effect of the supervision configuration on nuScenes~\cite{nuscenes}. Each method appears twice. Raw uses the standard target accumulated over 160 sweeps with $\lambda_{\text{grad}}=0$. Cleaned uses the filtered target of \S\ref{sec:impl} with $\lambda_{\text{grad}}=0.5$. Bold marks the better regime within each pair. The cleaned joint row is the main configuration.}
\label{tab:cleaning}
\setlength{\tabcolsep}{3pt}
\resizebox{\linewidth}{!}{%
\begin{tabular}{@{}llcccccccc@{}}
\toprule
 & & \multicolumn{2}{c}{0--50\,m} & \multicolumn{2}{c}{0--70\,m} & \multicolumn{4}{c}{0--80\,m} \\
\cmidrule(lr){3-4} \cmidrule(lr){5-6} \cmidrule(lr){7-10}
Method & Target & MAE (mm) $\downarrow$ & RMSE (mm) $\downarrow$ & MAE (mm) $\downarrow$ & RMSE (mm) $\downarrow$ & MAE (mm) $\downarrow$ & RMSE (mm) $\downarrow$ & iMAE (km$^{-1}$) $\downarrow$ & iRMSE (km$^{-1}$) $\downarrow$ \\
\midrule
\multirow{2}{*}{$\Dl$ only (horizon)}
 & Raw     & \textbf{911}  & 2895          & \textbf{1204} & 3908          & \textbf{1318} & 4393          & \textbf{4.38} & 13.43 \\
 & Cleaned & 964           & \textbf{2755} & 1251          & \textbf{3614} & 1352          & \textbf{3953} & 4.41          & \textbf{10.74} \\
\midrule
\multirow{2}{*}{$\mathbf{C}$ only (readout)}
 & Raw     & \textbf{913}  & 2914          & \textbf{1203} & 3970          & 1318          & 4471          & 4.54          & 14.87 \\
 & Cleaned & 956           & \textbf{2792} & 1227          & \textbf{3672} & \textbf{1317} & \textbf{4012} & \textbf{4.36} & \textbf{10.89} \\
\midrule
\multirow{2}{*}{$\Dl{+}\mathbf{C}$ (joint, main)}
 & Raw     & \textbf{900.7} & 2849.5          & 1204.5          & 3900.3          & 1311.9          & 4362.9          & \textbf{4.17} & 12.14 \\
 & Cleaned & 943.0          & \textbf{2783.9} & \textbf{1200.9} & \textbf{3610.7} & \textbf{1285.8} & \textbf{3920.2} & 4.22          & \textbf{10.85} \\
\midrule
\multirow{2}{*}{\FM{}}
 & Raw     & \textbf{1116.4} & \textbf{3329.5} & \textbf{1454.3} & 4482.8          & \textbf{1587.8} & 4992.7          & 5.44          & 15.43 \\
 & Cleaned & 1179.9          & 3337.2          & 1483.3          & \textbf{4282.5} & 1590.5          & \textbf{4640.0} & \textbf{5.34} & \textbf{12.90} \\
\bottomrule
\end{tabular}}
\end{table}

The cleaned configuration with gradient matching lowers every RMSE metric for every model relative to the raw configuration without it. RMSE@$80$ falls by 10.0 to 10.3\% for the RMS variants and by 7.1\% for \FM{}. iRMSE@$80$ falls by 10.6 to 26.8\%. Among the RMS variants, the largest drops occur for the single pathway variants, whose raw iRMSE@$80$ was highest ($14.87$ and $13.43$). MAE moves much less. The cleaned configuration improves the joint model by 2.0\% within 80\,m, leaves the readout variant and \FM{} unchanged within 0.2\%, and costs the horizon variant 2.6\%. The removed pixels are few but carry large errors under raw supervision. Squared metrics weight them heavily, while MAE is dominated by the many pixels that were already well supervised. The gap between \FM{} and the joint model is 17.4\% MAE@$80$ with the raw target and 19.2\% with the cleaned target, so the supervision configuration does not explain the architectural gain.

\subsection{Radar corruption protocol}
\label{app:corruption}

Corruption is applied at test time to the radar input of each nuScenes validation frame. The random generator for frame $i$ is seeded with $i$, so every model sees identical corrupted inputs. Spurious returns are added at uniformly random image positions with depths drawn uniformly from $[0.5, 120]$\,m. Their number is a percentage of the true returns in the frame. Removal keeps a random subset of the true returns, and the removed setting passes no radar at all. Depth noise adds zero mean Gaussian noise to the depth of each true return. We evaluate the SemoDepth and \FM{} models of Table~\ref{tab:conditions}, both selected by the rule of \S\ref{sec:impl}. Their clean results match Table~\ref{tab:main} within seed variation. Table~\ref{tab:corruption} lists every setting.

\begin{table}[ht]
\centering
\small
\setlength{\tabcolsep}{4pt}
\caption{Radar corruption at test time on nuScenes~\cite{nuscenes}, within 80\,m. Relative changes are with respect to the clean input of the same model.}
\label{tab:corruption}
\begin{tabular}{@{}lcccccc@{}}
\toprule
 & \multicolumn{3}{c}{SemoDepth} & \multicolumn{3}{c}{\FM{}} \\
\cmidrule(lr){2-4} \cmidrule(lr){5-7}
Setting & MAE (mm) & Change & RMSE (mm) & MAE (mm) & Change & RMSE (mm) \\
\midrule
Clean                        & 1284.5 & --         & 3950.6 & 1590.5 & --           & 4640.0 \\
\midrule
Spurious, $+10\%$            & 1285.7 & $+0.1\%$   & 3952.3 & 1616.9 & $+1.7\%$     & 4678.0 \\
Spurious, $+50\%$            & 1290.0 & $+0.4\%$   & 3957.9 & 1719.4 & $+8.1\%$     & 4829.8 \\
Spurious, $+100\%$           & 1294.8 & $+0.8\%$   & 3965.2 & 1858.4 & $+16.8\%$    & 5051.9 \\
Spurious, $+200\%$           & 1302.2 & $+1.4\%$   & 3974.5 & 2125.6 & $+33.6\%$    & 5481.9 \\
\midrule
Keep $75\%$                  & 1285.2 & $+0.1\%$   & 3951.9 & 1611.7 & $+1.3\%$     & 4678.1 \\
Keep $50\%$                  & 1286.8 & $+0.2\%$   & 3955.9 & 1660.1 & $+4.4\%$     & 4766.6 \\
Keep $25\%$                  & 1290.4 & $+0.5\%$   & 3965.9 & 1801.6 & $+13.3\%$    & 5032.7 \\
Keep $10\%$                  & 1296.8 & $+1.0\%$   & 3983.9 & 2239.7 & $+40.8\%$    & 5864.9 \\
Radar removed                & 1327.4 & $+3.3\%$   & 4071.9 & 4055.9 & $+155.0\%$   & 8935.4 \\
\midrule
Depth noise, $0.5$\,m        & 1284.5 & $0.0\%$    & 3950.6 & 1590.5 & $0.0\%$      & 4640.0 \\
Depth noise, $1$\,m          & 1284.5 & $0.0\%$    & 3950.6 & 1590.6 & $0.0\%$      & 4639.9 \\
Depth noise, $2$\,m          & 1284.6 & $0.0\%$    & 3950.6 & 1590.8 & $0.0\%$      & 4639.7 \\
Depth noise, $5$\,m          & 1284.6 & $0.0\%$    & 3950.8 & 1592.9 & $+0.2\%$     & 4639.9 \\
\bottomrule
\end{tabular}
\end{table}

Both models barely react to depth noise. They respond to whether and where returns appear, not to small changes in their depth. This matches the GSE input, which normalises each return by $d_{\max}$ and encodes it jointly with its image position. Removing radar from SemoDepth gives $1327$\,mm, close to a model trained without radar in the scan ($1306$\,mm, Table~\ref{tab:ablation}). The image pathway of SemoDepth works on its own, as the construction intends.

\subsection{Error by depth}
\label{app:rangebins}

\begin{figure}[ht]
\centering
\includegraphics[width=\linewidth]{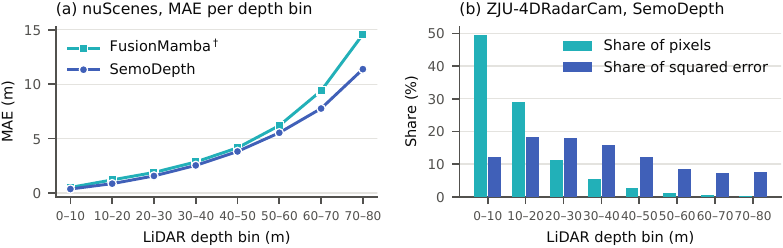}
\caption{Error by ground truth depth. (a) MAE per 10\,m bin on nuScenes for SemoDepth and \FM{}. (b) Share of pixels and share of squared error per bin for SemoDepth on ZJU-4DRadarCam.}
\label{fig:rangebins}
\end{figure}

Figure~\ref{fig:rangebins} and Table~\ref{tab:rangebins} break the errors down by ground truth depth, pooling all valid pixels of the evaluation split. On nuScenes SemoDepth has the lower MAE in every 10\,m bin. On both datasets the error grows with depth, and far pixels are under predicted. Pixels beyond 50\,m make up $5.1\%$ of the nuScenes pixels and $44.9\%$ of the squared error of SemoDepth. On ZJU-4DRadarCam, pixels beyond 70\,m are $0.4\%$ of the pixels and $7.5\%$ of the squared error, with a bias of $-7.3$\,m. MAE and RMSE are governed by nearly disjoint sets of pixels. This explains why SemoDepth can lead on MAE while TacoDepth keeps a small RMSE edge at 70 and 80\,m on ZJU-4DRadarCam.

\begin{table}[ht]
\centering
\small
\setlength{\tabcolsep}{3pt}
\caption{Error by ground truth depth. Pixel and SSE give the share of pixels and of the squared error in each bin. Errors and bias (prediction minus ground truth) are in metres. P50 and P99 are percentiles of the absolute error.}
\label{tab:rangebins}
\resizebox{\linewidth}{!}{%
\begin{tabular}{@{}lccccc|cccccc@{}}
\toprule
 & \multicolumn{5}{c|}{nuScenes} & \multicolumn{6}{c}{ZJU-4DRadarCam (SemoDepth)} \\
\cmidrule(lr){2-6} \cmidrule(lr){7-12}
Bin (m) & Pixel (\%) & MAE ours & MAE \FM{} & Bias ours & SSE ours (\%) & Pixel (\%) & MAE & P50 & P99 & Bias & SSE (\%) \\
\midrule
0--10  & 46.3 & 0.37  & 0.51  & 0.23   & 10.3 & 49.6 & 0.51 & 0.12 & 5.9  & 0.19  & 12.1 \\
10--20 & 25.1 & 0.86  & 1.21  & 0.28   & 13.3 & 29.0 & 1.08 & 0.29 & 10.9 & $-0.06$ & 18.4 \\
20--30 & 11.9 & 1.57  & 1.89  & $-0.19$  & 8.7  & 11.2 & 2.00 & 0.58 & 17.3 & $-0.69$ & 18.1 \\
30--40 & 7.2  & 2.54  & 2.86  & $-0.94$  & 10.3 & 5.4  & 2.69 & 0.84 & 24.7 & $-1.50$ & 16.0 \\
40--50 & 4.4  & 3.82  & 4.16  & $-2.15$  & 12.4 & 2.6  & 3.45 & 1.18 & 32.2 & $-2.33$ & 12.1 \\
50--60 & 2.8  & 5.54  & 6.20  & $-3.92$  & 15.1 & 1.2  & 4.33 & 1.64 & 41.3 & $-3.31$ & 8.6 \\
60--70 & 1.6  & 7.77  & 9.39  & $-6.44$  & 15.6 & 0.7  & 5.31 & 2.04 & 49.6 & $-4.62$ & 7.1 \\
70--80 & 0.7  & 11.38 & 14.58 & $-10.56$ & 14.2 & 0.4  & 7.48 & 2.77 & 63.0 & $-7.34$ & 7.5 \\
\bottomrule
\end{tabular}}
\end{table}

\subsection{Failure cases}
\label{app:failures}

\begin{figure}[ht]
\centering
\includegraphics[width=\linewidth]{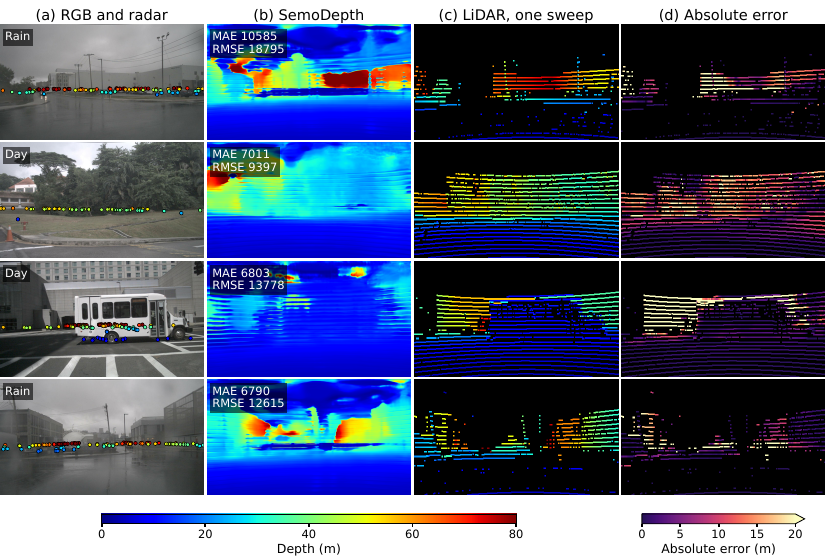}
\caption{Worst failure cases of SemoDepth on nuScenes~\cite{nuscenes}. The rows are the four frames with the highest MAE@$80$ among distinct scenes (ranks 1, 5, 7, and 8 of $5{,}869$). Column (a) shows the RGB input with radar returns coloured by depth. Column (b) shows the SemoDepth prediction with the MAE and RMSE of the frame in millimetres. Column (c) shows LiDAR ground truth from one sweep, and column (d) the absolute error at ground truth pixels, clipped at 20\,m. Depth colours are clipped at 80\,m.}
\label{fig:failures}
\end{figure}

Figure~\ref{fig:failures} shows where SemoDepth fails worst. We rank all $5{,}869$ validation frames by their MAE@$80$ and keep the four worst frames from distinct scenes. Frames ranked 2, 3, 4, and 6 belong to the scene of the worst frame, so the selection moves on to the next scene. No frame is chosen by hand. The median frame has an MAE@$80$ of $1078$\,mm, and the 95th percentile is $2738$\,mm. The worst $1\%$ of frames (59 frames, MAE@$80$ above $4070$\,mm) come from only 14 scenes. Rain is overrepresented among them, with 40 of the 59 frames against $18\%$ of all frames. Night accounts for 14 of them against $10\%$ of all frames. In these frames $44\%$ of the absolute error comes from pixels beyond 50\,m, against $28\%$ over all frames.

In both rain frames (rows 1 and 4) the image is grey and low in contrast. The prediction contains a band of near depth along the horizon just below the radar returns, and distant facades are placed too close. In row 1 a warehouse facade at 60 to 75\,m is split into a part at 30 to 50\,m and a part saturated at 80\,m. In row 2 a tree line that LiDAR and radar place at 40 to 55\,m is predicted at 25 to 35\,m. In row 3 the depth of a nearby bus spreads into the background beside and above it. In all four frames, more than $90\%$ of the pixels with errors above 20\,m are under predictions, which matches the far range bias in Table~\ref{tab:rangebins}.

\section{Additional results on ZJU-4DRadarCam}

\subsection{Inverse depth metrics and methods with a monocular prior}
\label{sec:zju_idepth}

Table~\ref{tab:zju_idepth} gives the inverse depth metrics for the comparison of Table~\ref{tab:zju}. SemoDepth has the best iMAE and iRMSE at every range, about $18\%$ and $8\%$ below TacoDepth. Table~\ref{tab:zju_tracks} adds methods that consume a monocular depth prior at test time. All methods on ZJU-4DRadarCam share one radar scan and one image per frame, the released interpolated training target, and sparse evaluation. SemoDepth beats every one of the five RadarCam-Depth configurations~\cite{RadarCam-Depth} on MAE and RMSE at every range. The other four configurations reach $1205.1$ to $1365.4$\,mm MAE@$80$. SemoDepth matches the best RadarCam-Depth configuration on iMAE and iRMSE. The version of TacoDepth that also uses a monocular prior is the strongest entry in the table.

\begin{table}[ht]
\centering
\small
\setlength{\tabcolsep}{4pt}
\caption{Inverse depth metrics on ZJU-4DRadarCam~\cite{RadarCam-Depth}, complementing Table~\ref{tab:zju}. Baseline numbers are as reported by TacoDepth~\cite{TacoDepth}. Best per column in bold.}
\label{tab:zju_idepth}
\resizebox{\linewidth}{!}{%
\begin{tabular}{@{}lcccccc@{}}
\toprule
\multicolumn{1}{c}{\multirow{2}{*}{Method}} & \multicolumn{2}{c}{0--50\,m} & \multicolumn{2}{c}{0--70\,m} & \multicolumn{2}{c}{0--80\,m} \\
\cmidrule(lr){2-3} \cmidrule(lr){4-5} \cmidrule(lr){6-7}
 & iMAE (km$^{-1}$) $\downarrow$ & iRMSE (km$^{-1}$) $\downarrow$ & iMAE (km$^{-1}$) $\downarrow$ & iRMSE (km$^{-1}$) $\downarrow$ & iMAE (km$^{-1}$) $\downarrow$ & iRMSE (km$^{-1}$) $\downarrow$ \\
\midrule
DORN~\cite{DORN}                     & 19.8          & 31.9          & 19.8          & 31.9          & 19.9          & 31.9          \\
Singh et al.~\cite{Singh}            & 18.1          & 35.3          & 18.0          & 35.2          & 17.9          & 35.1          \\
TacoDepth~\cite{TacoDepth}           & 12.8          & 25.0          & 12.7          & 24.9          & 12.7          & 24.9          \\
SemoDepth (Ours)                     & \textbf{10.4} & \textbf{22.8} & \textbf{10.4} & \textbf{22.8} & \textbf{10.4} & \textbf{22.8} \\
\bottomrule
\end{tabular}}
\end{table}

\begin{table}[ht]
\centering
\small
\setlength{\tabcolsep}{5pt}
\caption{ZJU-4DRadarCam~\cite{RadarCam-Depth} within 80\,m, grouped by whether a method uses a monocular depth prior. Numbers for other methods come from RadarCam-Depth~\cite{RadarCam-Depth} and TacoDepth~\cite{TacoDepth}. For RadarCam-Depth we list the best of its five configurations.}
\label{tab:zju_tracks}
\resizebox{\linewidth}{!}{%
\begin{tabular}{@{}llcccc@{}}
\toprule
Monocular prior & Method & MAE (mm) & RMSE (mm) & iMAE (km$^{-1}$) & iRMSE (km$^{-1}$) \\
\midrule
\multirow{4}{*}{No}
 & DORN~\cite{DORN}              & 2447.6 & 4760.0 & 19.9 & 31.9 \\
 & Singh et al.~\cite{Singh}     & 1979.5 & 4309.3 & 17.9 & 35.1 \\
 & TacoDepth~\cite{TacoDepth}    & 1201.1 & 2990.7 & 12.7 & 24.9 \\
 & SemoDepth (Ours)              & 1137.2 & 3053.0 & 10.4 & 22.8 \\
\midrule
\multirow{2}{*}{Yes}
 & RadarCam-Depth~\cite{RadarCam-Depth} & 1183.5 & 3229.0 & 10.4 & 22.8 \\
 & TacoDepth~\cite{TacoDepth}           & 1032.5 & 2850.3 & 9.4  & 20.9 \\
\bottomrule
\end{tabular}}
\end{table}

\subsection{Qualitative results}
\label{app:qualitative-zju}

\begin{figure}[ht]
\centering
\includegraphics[width=\linewidth]{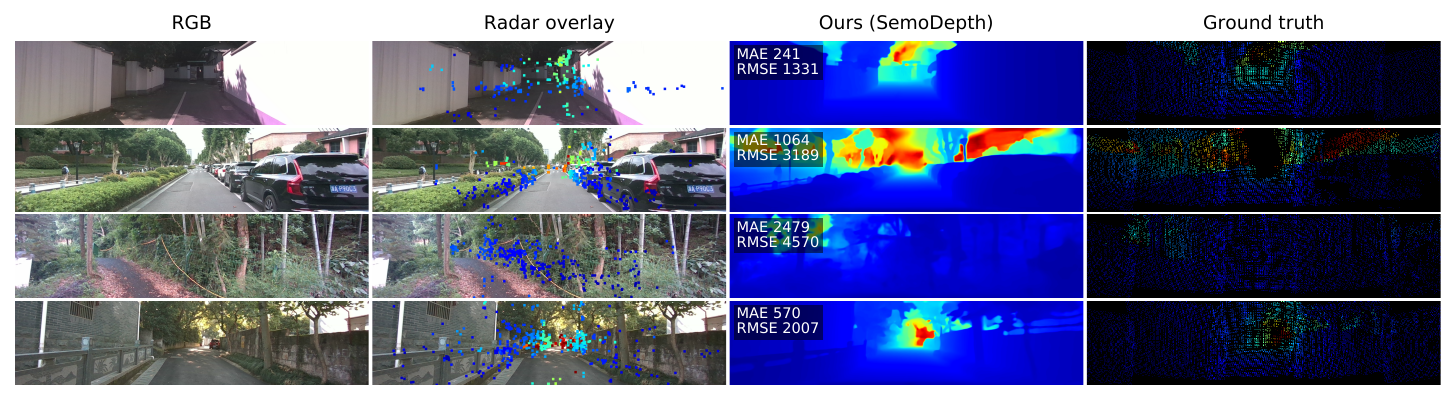}
\caption{Qualitative results on ZJU-4DRadarCam~\cite{RadarCam-Depth}. Columns show the RGB input, radar returns over the RGB, the SemoDepth prediction, and LiDAR ground truth from a single sweep. Each prediction panel gives the MAE and RMSE of the frame in millimetres within 80\,m, and the colour map is clipped at 80\,m. The frames were drawn at random from the test split (seed 0, indices 1105, 2093, 2607, and 3481 of 4{,}097), so the spread of errors reflects the test set rather than selection. SemoDepth recovers vehicle and lane geometry although the radar provides only about 380 returns per frame.}
\label{fig:qualitative-zju}
\end{figure}

Figure~\ref{fig:qualitative-zju} shows SemoDepth on four test frames of ZJU-4DRadarCam~\cite{RadarCam-Depth}, together with the radar input and the LiDAR ground truth. No public ZJU weights exist for DORN~\cite{DORN}, Singh et al.~\cite{Singh}, or TacoDepth~\cite{TacoDepth}, so the figure compares against ground truth.

\section{Extended limitations}
\label{app:limitations}

The main text lists the limits that most affect our claims. This section covers them in more detail.

Supervision noise. The nuScenes target accumulates LiDAR over 160 sweeps, so moving objects leave trails and the far field contains artefacts. Our filter on the upper image removes the worst of them but not all. Table~\ref{tab:cleaning} shows that the cleaned configuration, which also adds gradient matching, mainly lowers RMSE and leaves MAE almost unchanged for both architectures. Some artefacts remain visible in qualitative depth maps. A principled treatment of dynamic objects in the accumulated target remains open.

Fairness of baseline comparisons. Our protocol matches the published protocols of TacoDepth and RadarCam-Depth in split, input resolution, accumulation window (160 LiDAR sweeps at 20\,Hz, about 8\,s), sparse evaluation, and learning rate schedule. On ZJU-4DRadarCam every method uses the same released training target. Three differences remain. Our loss uses log space, Huber, and gradient terms instead of the plain L1 terms of TacoDepth. Our ZJU runs train for up to 100 epochs instead of 50. Our nuScenes target is filtered. The loss and the filter apply identically to every SemoDepth variant and to \FM{}, so they cannot explain the gap between the two architectures. Published baselines enter through their reported numbers. TacoDepth has no public release and cannot be retrained. \FM{} serves as the controlled baseline. Single seed ablations carry seed variance of the order of the smaller gaps between rows.

Radar corruption and missing radar. Spurious, sparse, and missing radar are measured in \S\ref{sec:robustness} and Table~\ref{tab:corruption}. SemoDepth degrades gracefully in each case. We did not test a systematic bias in radar depth, such as a calibration offset, or simultaneous failures of camera and radar, such as heavy spray.

Far tail on ZJU-4DRadarCam. SemoDepth under predicts far depth on both datasets (Table~\ref{tab:rangebins}). On ZJU-4DRadarCam the bias reaches $-7.3$\,m between 70 and 80\,m, where 4D radar is sparsest, and this small tail explains the RMSE gap to TacoDepth at 70 and 80\,m. Sparse far supervision and the compression of log space are likely causes.

Margin and radar density. The margin over TacoDepth is 5 to 8\% MAE on ZJU-4DRadarCam and about 30\% on nuScenes. ZJU uses 4D radar with elevation and about 380 returns per frame, roughly ten times denser than the 40 to 100 returns of nuScenes. Denser and less ambiguous radar leaves less work for any fusion design and less room for selection to add value. We have not tested a 3D radar dataset of comparable scale to nuScenes.

Night and weather. At night, fusion before the scan matches or beats SemoDepth on linear metrics, and the outcome depends on the seed (Table~\ref{tab:conditions}). Rain is covered, but fog is absent from nuScenes and untested.

Embedded deployment. All latencies are measured on a workstation GPU at FP32 and batch 1. We have not measured latency, memory, or the behaviour of the selective scan kernel on embedded automotive hardware, and quantisation is untested.

Safety of dense depth failures. Depth errors concentrate at far range, at night, and in the failure cases of Figure~\ref{fig:failures}. A planner that trusts dense depth uniformly could act on these errors. Any deployment needs evaluation by scene type, uncertainty estimates, and monitoring of inputs that differ from the training data.

Scope and statistics. We evaluate front camera and automotive radar only. The argument for selection inside the scan does not depend on the sensors, but we have not tested LiDAR and camera, radar and thermal, or other pairs. Table~\ref{tab:main} and the blend comparison use three seeds. Other ablations, the condition splits, and the corruption study use one seed, as is common in this field~\cite{RadarCam-Depth,TacoDepth,CaFNet,LiSBD}.

\section{Broader impacts}
\label{sec:broader}

Robust, low cost, and weather tolerant metric depth can enable safer driver assistance on vehicles that cannot afford automotive LiDAR. SemoDepth relies on radar already common in vehicles and degrades gracefully when radar is poor. It could lower the cost of features such as forward collision warning, automatic emergency braking, and adaptive cruise control. These features help most in cost sensitive markets and in weather where camera only systems degrade.

Better scene understanding from radar and camera is also dual use. The same depth predictions could feed surveillance or military systems. Large scale deployment shares the usual risks of automated perception. Failures can concentrate on rare scenes or atypical objects, such as pedestrians with unusual shapes or clothing. Such failures can have safety consequences if a planner assumes uniform accuracy. We do not study geographic or demographic robustness. nuScenes and ZJU-4DRadarCam come from narrow regions (Boston and Singapore, and one campus), which limits how far our numbers transfer.

The released checkpoints are research artefacts and not production components for driver assistance. They should be used only for research, and any deployment should add evaluation by scene type and monitoring for unfamiliar inputs. The method adds no capability for harm beyond existing radar camera depth methods~\cite{CaFNet,TacoDepth,LiSBD}, so we do not restrict access.

\section{Compute resources}
\label{sec:compute}

All experiments run on one workstation with one NVIDIA RTX PRO 6000 Blackwell Workstation Edition GPU (96\,GB, CUDA 13.0). Training uses bf16 mixed precision (\S\ref{sec:impl}) and needs about 24\,GB at batch size 12. A full 50 epoch run on nuScenes ($27{,}430$ training frames) takes about 26 hours. A ZJU-4DRadarCam run ($26{,}055$ training frames, 100 epochs) takes about 24 hours. The filtering of the supervision target is a single pass over the training and validation data, about 40 minutes on 40 CPU threads. Inference takes $26.8$\,ms per frame at $900{\times}1600$.

The configurations of Tables~\ref{tab:main}, \ref{tab:zju}, \ref{tab:cleaning}, and \ref{tab:daynight} took about 13 GPU days. The additional ablations, the corruption study, and the analyses by condition and depth took about 3 GPU days. Including preliminary experiments that are not reported, the project used about 28 GPU days on the same workstation.

\section{Code and reproducibility}
\label{sec:code}

Code, preprocessing scripts, and trained checkpoints for SemoDepth are available at \url{https://github.com/HouV001/SemoDepth}. The repository contains the model (\texttt{mambadepth\_fusion/}), training scripts for both datasets, an evaluation script for nuScenes, the LiDAR accumulation utility, the filtering of the supervision target, and unit tests. The tests include the check that numerically verifies Eq.~\eqref{eq:init-parity} on a CUDA device. The README describes the environment and the data preparation. Training scripts fix the random seeds. Checkpoint selection on the validation split happens automatically during training.

\section{Licenses for existing assets}
\label{sec:licenses}

Table~\ref{tab:licenses} lists the external datasets, code, and pretrained weights we use, with versions, licenses, and roles. All assets are used under their licenses. nuScenes~\cite{nuscenes} is used under its non commercial research license, and no asset is redistributed.

\begin{table}[ht]
\centering
\small
\caption{External assets used in SemoDepth. Versions are the snapshots used for our experiments. Licenses are summarised, and the full texts ship with each asset.}
\label{tab:licenses}
\resizebox{\linewidth}{!}{%
\begin{tabular}{@{}llll@{}}
\toprule
Asset & Version & License & Role in this work \\
\midrule
nuScenes~\cite{nuscenes} & v1.0-trainval & CC BY-NC-SA 4.0 & Primary benchmark \\
ZJU-4DRadarCam~\cite{RadarCam-Depth} & official release & per upstream repository & Secondary benchmark \\
nuscenes-devkit & 1.1.x & Apache 2.0 & LiDAR accumulation and projection \\
ResNet-34 (torchvision) & ImageNet1K-V1 & BSD-3-Clause & Image encoder initialisation \\
\texttt{mamba-ssm} & 1.2.x & Apache 2.0 & Selective scan CUDA kernel \\
\texttt{causal-conv1d} & 1.2.x & BSD-3-Clause & Mamba dependency \\
PyTorch & 2.x & BSD-3-Clause & Training framework \\
PCA-GM~\cite{pcagm} & reference implementation & per upstream repository & Building block of the radar GSE \\
TacoDepth~\cite{TacoDepth} & paper formulae & no code reused & Baseline and loss reference \\
RadarCam-Depth~\cite{RadarCam-Depth} & ZJU release & per upstream repository & Baseline and host of the ZJU dataset \\
\bottomrule
\end{tabular}}
\end{table}

Each method we compare against is cited where it is used. Our code consists of new training, evaluation, and visualisation scripts that use the listed assets through their public interfaces.

\end{document}